\documentclass{article} 
\usepackage{iclr2023_conference,times}

\usepackage{amsmath,amsfonts,bm}

\def\eqref#1{equation~\ref{#1}}

\def\1{\bm{1}}

\DeclareMathAlphabet{\mathsfit}{\encodingdefault}{\sfdefault}{m}{sl}
\SetMathAlphabet{\mathsfit}{bold}{\encodingdefault}{\sfdefault}{bx}{n}

\DeclareMathOperator*{\argmin}{arg\,min}

\usepackage{url}

\usepackage{multirow}
\usepackage{multicol}
\usepackage{microtype}
\usepackage{graphicx}
\usepackage{subfig}
\usepackage{booktabs} 
\usepackage{caption}
\usepackage{wrapfig}

\usepackage[hidelinks]{hyperref}

\usepackage{algorithm}
\usepackage{algorithmic}
\usepackage{amsmath}
\usepackage{amssymb}
\usepackage{mathtools}
\usepackage{amsthm}
\usepackage{physics}

\usepackage[capitalize,noabbrev]{cleveref}

\def\circlearrow{\ensuremath{%
  \rotatebox[origin=c]{100}{$\circlearrowright$}}}

\newcommand{\bZero}{\mathbf{0}}

\newcommand{\bff}{\mathbf{f}}

\newcommand{\bs}{\mathbf{s}}

\newcommand{\bw}{\mathbf{w}}
\newcommand{\bx}{\mathbf{x}}

\newcommand{\bz}{\mathbf{z}}

\newcommand{\bI}{\mathbf{I}}

\newcommand{\bepsilon}{\boldsymbol{\epsilon}}

\Crefname{equation}{Eq.}{Eqs.}
\Crefname{figure}{Fig.}{Figs.}
\Crefname{tabular}{Tab.}{Tabs.}
\Crefname{algorithm}{Alg.}{Algs.}

\theoremstyle{plain}
\newtheorem{problem}{Problem}
\newtheorem{theorem}{Theorem}[section]
\newtheorem{proposition}[theorem]{Proposition}

\theoremstyle{definition}

\theoremstyle{remark}

\newenvironment{proofsketch}{\proof}{\endproof}

\usepackage[textsize=tiny]{todonotes}
\newcommand{\revision}[1]{#1}

\newcommand{\se}[1]{}
\newcommand{\lingxiao}[1]{}
\newcommand{\js}[1]{}
\newcommand{\chenlin}[1]{}

\newcommand{\model}[1]{{Dual Diffusion Implicit Bridges}}
\newcommand{\modellower}[1]{{dual diffusion implicit bridges}}
\newcommand{\modelshort}[1]{{DDIBs}}

\title{\model{} for \\Image-to-Image Translation}

\author{Xuan Su$^1$ \qquad Jiaming Song$^2$ \qquad Chenlin Meng$^1$ \qquad Stefano Ermon$^{1,3}$\\
$^1$Stanford University \qquad $^2$NVIDIA \qquad $^3$CZ Biohub\\
\texttt{ \{suxuan,chenlin,ermon\}@cs.stanford.edu,jiamings@nvidia.com}
}

\iclrfinalcopy 
\begin{document}

\maketitle

\begin{abstract}
Common image-to-image translation methods rely on joint training over data from both source and target domains. The training process requires concurrent access to both datasets, which hinders data separation and privacy protection; and existing models cannot be easily adapted for translation of new domain pairs.
\se{privacy argument here will not be understandable i think. maybe say each pair requires training from scratch and access to paired data, which might be scarce or even unavailable due to privacy.. }
We present \model{} (\modelshort{}), an image translation method based on diffusion models, that circumvents training on domain pairs. Image translation with \modelshort{} relies on two diffusion models trained independently on each domain, and is a two-step process: \modelshort{} first obtain latent encodings for source images with the source diffusion model, and then decode such encodings using the target model to construct target images. Both steps are defined via \revision{ordinary differential equations (ODEs)}, thus the process is cycle consistent only up to discretization errors of the ODE solvers. Theoretically, we interpret \modelshort{} as concatenation of source to latent, and latent to target Schr\"odinger Bridges, a form of entropy-regularized optimal transport, to explain the efficacy of the method. Experimentally, we apply \modelshort{} on synthetic and high-resolution image datasets, to demonstrate their utility in a wide variety of translation tasks and their inherent optimal transport properties.\se{not clear what demonstrating the connection experimentally means. it might make sense to explicitly say why this solution addresses the motivating problems}
\end{abstract}
\section{Introduction}
\label{chap:introduction}
Transferring images from one domain to another while preserving the content representation is an important problem in computer vision, with wide applications that span style transfer \citep{xu2021drb,sinha2021d2c} and semantic segmentation \citep{li2020simplified}. \se{inpainting might not be the best motivation for your story. same for colorization actually, because it's easy to remove color}In tasks such as style transfer, it is usually difficult to obtain paired images of realistic scenes and their artistic renditions. Consequently, unpaired translation methods are particularly relevant, since only the datasets, and not the one-to-one correspondence between image translation pairs, are required. \se{be more explicit about paired vs unpaired.}Common methods on unpaired translation are based on generative adversarial networks (GANs, \cite{goodfellow2014generative,zhu2017unpaired}) or normalizing flows \citep{grover2020alignflow}. Training such models typically involves minimizing an adversarial loss between a specific pair of source and target datasets.

While capable of producing high-quality images, these methods suffer from a severe drawback in their \emph{adaptability} to alternative domains. Concretely, a translation model on a source-target pair is trained specifically for this domain pair.\se{what's the difference between tuning and training?} Provided a different pair, existing, bespoke models cannot be easily adapted for translation. If we were to do pairwise translation among a set of domains, the total number of models needed is quadratic in the number of domains -- an unacceptable computational cost in practice. One alternative is to find a shared domain that connects to each source / target domains as in StarGANs \citep{choi2018stargan}. However, the shared domain needs to be carefully chosen \textit{a priori}; if the shared domain contains less information than the target domain (\textit{e.g.} sketches v.s. photos), then it creates an unwanted information bottleneck between the source and target domains.

An additional disadvantage of existing models resides in their \emph{lack of privacy protection} of the datasets: training a translation model requires access to both datasets simultaneously. Such setting may be inconvenient or impossible, when data providers are reluctant about giving away their data; or for certain privacy-sensitive applications such as medical imaging. For example, quotidian hospital usage may require translation of patients' X-ray and MRI images taken from machines in other hospitals. Most existing methods will fail in such scenarios, as joint training requires aggregating confidential imaging data across hospitals, which may violate patients' privacy.\se{this paragraph needs rewording. people might wonder how the task is even possible if there is no data from one of the domains}

\begin{figure*}
\vspace{-.6cm}
\begin{center}
\centerline{\includegraphics[width=\linewidth]{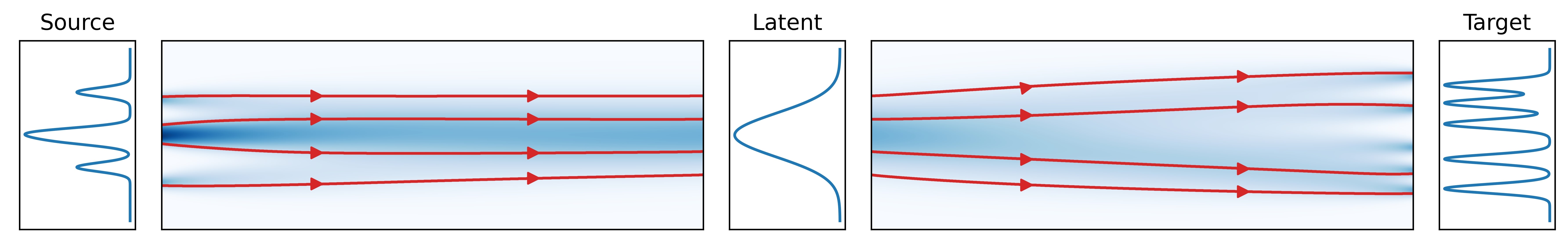}}
\vspace{-.1cm}
\centerline{\includegraphics[width=1.06\linewidth]{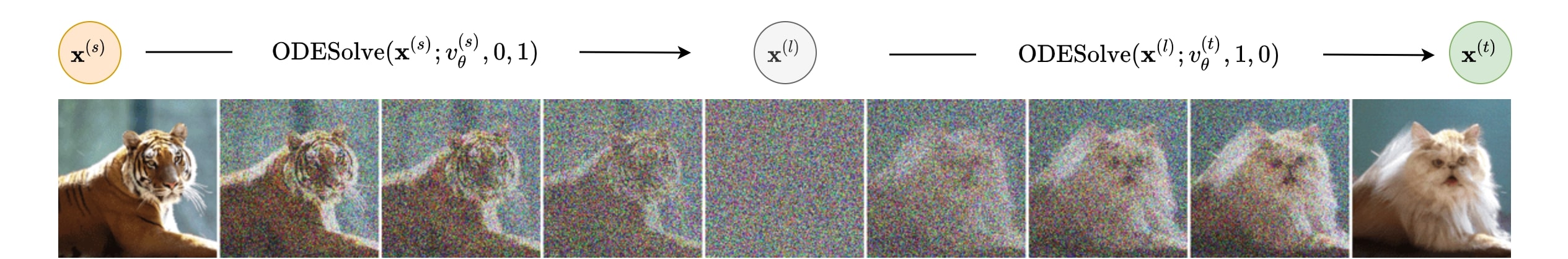}}
\vspace{-.1cm}
\caption{\textbf{\model{}}: \modelshort{} leverage two ODEs for image translation. Given a source image $\bx^{(s)}$, the source ODE runs in the forward direction to convert it to the latent $\bx^{(l)}$, while the target, reverse ODE then constructs the target image $\bx^{(t)}$. (\textit{Top}) Illustration of the \modelshort{} idea between two one-dimensional distributions. (\textit{Bottom}) \modelshort{} from a tiger to a cat using a pretrained conditional diffusion model.}
\vspace{-.8cm}
\label{fig:figure_1}
\end{center}
\end{figure*}

In this paper, we seek to mitigate both problems of existing image translation methods. We present \model{} (\modelshort{}), an image-to-image translation method inspired by recent advances in diffusion models~\citep{song2020denoising,song2020score}, that decouples paired training, and empowers the domain-specific diffusion models to stay applicable in other pairs wherever the domain appears again as the source or the target. \se{odd wording, rephrase. Xuan: changes reappears}Since the training process now concentrates on one dataset at a time, \modelshort{} can also be applied in federated settings, and not assume access to both datasets during model training. As a result, owners of domain data can effectively preserve their data privacy.\se{federated learning here comes out of nowhere. need to justify why somehting like cyclegan could not be trained in a federated way}

Specifically, \modelshort{} are developed based on the method known as denoising diffusion implicit models (DDIMs, \cite{song2020denoising}). DDIMs invent a particular parameterization of the diffusion process, that creates a smooth, deterministic and reversible mapping between images and their latent representations. This mapping is captured using the solution to a so-called \emph{probability flow} (PF) ordinary differential equation (ODE) that forms the cornerstone of \modelshort{}. Translation with \modelshort{} on a source-target pair requires two different PF ODEs: the source PF ODE converts input images to the latent space; while the target ODE then synthesizes images in the target domain.

Crucially, trained diffusion models are specific to the individual domains, and rely on no domain pairing information. Effectively, \modelshort{} make it possible to save a trained model of a certain domain for future use, when it arises as the source or target in a new pair. Pairwise translation with \modelshort{} requires only a \emph{linear} number of diffusion models (which can be further reduced with conditional models \citep{dhariwal2021diffusion}), and training does not require scanning both datasets concurrently.
\se{not sure i buy this, isn't the latent space the shared domain as in stargan essentially?}

Theoretically, we analyze the \modelshort{} translation process to highlight two important theoretical properties. First, the probability flow ODEs in \modelshort{}, in essence, comprise the solution of a special Schr\"odinger Bridge Problem (SBP) with linear or degenerate drift \citep{chen2021likelihood}, between the data and the latent distributions. This justification of \modelshort{} from an optimal transport viewpoint that alternative translation methods lack serves as a theoretical advantage of our method, as \modelshort{} are the most OT-efficient translation procedure while alternate methods may not be.\se{why?}
Second, \modelshort{} guarantee exact cycle consistency: translating an image to and back from the target space reinstates the original image, only up to discretization errors introduced in the ODE solvers.\se{define this}\se{not sure what's the takeaway/significance of this paragraph. might need reprhasing to highlight the main takeaway}

Experimentally, we first present synthetic experiments on two-dimensional datasets to demonstrate \modelshort{}' cycle-consistency property. We then evaluate our method on a variety of image modalities, with qualitative and quantitative results: we validate its usage in example-guided color transfer, paired image translation, and conditional ImageNet translation. These results establish \modelshort{} as a scalable, theoretically rigorous addition to the family of unpaired image translation methods.

\vspace{-.2cm}
\section{Preliminaries}
\label{sec:preliminaries}
\vspace{-.1cm}
\subsection{Score-based Generative Models (SGMs)}
\vspace{-.1cm}
While our actual implementation utilizes DDIMs, we first briefly introduce the broader family of models known as score-based generative models. Two representative models of this family are \emph{score matching with Langevin dynamics} (SMLD) \citep{song2019generative} and \emph{denoising diffusion probabilistic models} (DDPMs) \citep{ho2020denoising}.
Both methods are contained within the framework of Stochastic Differential Equations (SDEs) proposed in \cite{song2020score}.

\paragraph{Stochastic Differential Equation (SDE) Representation} \cite{song2020score,anderson1982reverse} use a forward and a corresponding backward SDE to describe general diffusion and the reversed, generative processes:
\begin{align}
    \dd{\bx} = \bff(\bx, t) \dd{t} + g(t) \dd{\bw}, \quad \dd{\bx} = [\bff - g^{2} \nabla_{\bx} \log p_t(\bx)] \dd{t} + g(t) \dd{\bw} \label{eq:sgm_sde}
\end{align}
where $\bw$ is the standard Wiener process, $\bff(\bx, t)$ is the vector-valued drift coefficient, $g(t)$ is the scalar diffusion coefficient, and $\nabla_{\bx} \log p_t(\bx)$ is the score function of the noise perturbed data distribution (as defined by the forward SDE with initial condition $p_0(\bx)$ being the data distribution). At the endpoints $t = \{ 0, 1 \}$, the forward \Cref{eq:sgm_sde} admits the data distribution $p_0$ and the easy-to-sample prior $p_1$ as the boundary distributions.
Within this framework, the SMLD method can be described using a \emph{Variance-Exploding} (VE) SDE with increasing noise scales $\sigma(t)$: $\dd{\bx} = \sqrt{\dd{[\sigma^{2}(t)]} / \dd{t}} \dd{\bw}$. In comparison, DDPMs are endowed with a \emph{Variance-Preserving} (VP) SDE: $\dd{\bx} = -[\beta(t) / 2] \bx \dd{t} + \sqrt{\beta(t)} \dd{\bw}$ with $\beta(t)$ being another noise sequence. Notably, the VP SDE can be reparameterized into an equivalent VE SDE \citep{song2020denoising}.

\paragraph{Probability Flow ODE} Any diffusion process can be represented by a \emph{deterministic} ODE that carries the same marginal densities as the diffusion process throughout its trajectory. This ODE is termed the \emph{probability flow} (PF) ODE \citep{song2020score}. PF ODEs enable \emph{uniquely identifiable encodings} \citep{song2020score} of data, and are central to \modelshort{} as we solve these ODEs for forward and reverse conversion between data and their latents. \lingxiao{DDIB not yet introduced in this section? Also what's latent? prior?}For the forward SDE introduced in \Cref{eq:sgm_sde}, the equivalent PF ODE \lingxiao{Is this equivalent to forward or backward SDE? And what about the boundary conditions?}holds the following form:
\begin{align}
    \dd{\bx} = \left[ \bff(\bx, t) - \frac{1}{2} g(t)^{2} \nabla_{\bx} \log p_t(\bx) \right] \dd{t}
    \label{eq:pfode_sgm}
\end{align}
which follows immediately from the SDEs given the score function. In practice, we use $\theta$-parameterized score networks $\bs_{t,\theta} \approx \nabla_{\bx} \log p_t(\bx)$ to approximate the score function. \revision{Training such networks relies on a variational lower bound, described in \cite{ho2020denoising} and in \Cref{appendix:ddim_details}.}
We may then employ numerical ODE solvers to solve the above ODE and construct $\bx$ at different times. Empirically, it has been demonstrated that SGMs have relatively low discretization errors when reconstructing $\bx$ at $t = 0$ via ODE solvers~\citep{song2020denoising}. For conciseness, we use $v_\theta = \dd{\bx} / \dd{t}$ to denote the $\theta$-parameterized velocity field (as defined from \Cref{eq:pfode_sgm}, where we replace $\nabla_{\bx} \log p_t(\bx)$ with $\bs_{t,\theta}$), and use the symbol $\mathrm{ODESolve}$ to denote the mapping from $\bx(t_0)$ to $\bx(t_1)$:
\begin{align}
    \mathrm{ODESolve}(\bx(t_0); v_\theta, t_0, t_1) = \bx(t_0) + \int_{t_0}^{t_1} v_\theta(t, \bx(t)) \dd{t}, \label{eq:odesolve}
\end{align}
which allows us to abstract away the exact model (be it a score-based or a diffusion model), or the integrator used.
In our experiments, we implement the ODE solver in DDIMs \citep{song2020denoising} (\Cref{appendix:ddim_details})\lingxiao{Again should probably introduce DDIM formally in this section first}; while we acknowledge other available ODE solvers that are usable within our framework, such as the DPM-solver \citep{lu2022dpm}, the Exponential Integrator \citep{zhang2022fast}, and the second-order Heun solver~\citep{karras2022elucidating}.\lingxiao{The sentence before the semicolon does not mention the particular solver/integrator of the ODE, but the second half does?}

\subsection{Schr\"odinger Bridge Problem (SBP)}

Our analysis shows that \modelshort{} are Schr\"odinger Bridges \citep{chen2016relation,leonard2013survey} between distributions. Let $\Omega = C([0, 1]; \mathbb{R}^{n})$ be the path space of $\mathbb{R}^{n}$-valued continuous functions over the time interval $[0, 1]$; and $\mathcal{D}(p_{0}, p_{1})$ be the set of distributions over $\Omega$
\se{what exactly is this object?}, with marginals $p_{0}$, $p_{1}$ at time $t=0$, $t=1$, respectively. Given a prior reference measure $W$\footnote{In our application, the reference measure is set to the measure of \Cref{eq:sgm_sde}, as per \cite{chen2021likelihood}.}, the well-known \emph{Schr\"odinger Bridge Problem} (SBP) seeks the most probable evolution across time $t$ between the marginals $p_{0}$ and $p_{1}$:
\begin{problem}[Schr\"odinger Bridge Problem] With prescribed distributions $p_{0}, p_{1}$ and a reference measure $W$ as the prior, the SBP finds a distribution from $\mathcal{D}(p_{0}, p_{1})$ that minimizes its KL-divergence to $W$: $P_{\text{SBP}} \coloneqq \argmin \{ D_{\text{KL}}(P \| W) \mid P \in \mathcal{D}(p_{0}, p_{1}) \}$.\js{argmin over $P$ and defined over all valid couplings}
\lingxiao{This sentence seems a bit inaccurate without specifying the initial distribution. In usual SBP, $W$ is the Brownian motion with the volume measure as the initial distribution.}
\label{problem:sbp}
\end{problem}
\se{this definition is missing too many details. what is W? what is the set you optimize over? at the very least we need to rule out a trivial "discontinuous" solution that has the right endpoints and W for every other t}
The minimizer, $P_{\text{SBP}}$, is dubbed the \emph{Schr\"odinger Bridge} between $p_{0}$ and $p_{1}$ over prior $W$. The SBP has connections to the Monge-Kantorovich (MK) optimal transport problem \citep{chen2021stochastic}. While the basic MK problem seeks the cost-minimizing plan to transport masses between distributions, the SBP incorporates an additional entropy term (for details, see Page 61 of \cite{peyre2019computational}) \lingxiao{The ``latter", i.e., the regularized OT, also has an entropy term}.

\paragraph{Relationship Between SBPs and SGMs} \cite{chen2021likelihood} establishes\se{what's a strong vs weak connection? avoid subjective statements} connections between SGMs and SBPs. In summary, SGMs are implicit optimal transport models, corresponding to SBPs with linear or degenerate drifts. General SBPs additionally accept fully nonlinear diffusion. To formalize this observation, the authors first establish similar forward and backward SDEs for SBPs:
\begin{align}
    \dd{\bx} = [ \bff + g^{2} \nabla_{\bx} \log \Phi_{t}(\bx)] \dd{t} + g(t) \dd{\bw}, \quad \dd{\bx} = [ \bff - g^{2} \nabla_{\bx} \log \hat{\Phi}_{t}(\bx)] \dd{t} + g(t) \dd{\bw} \label{eq:sbp_sde}
\end{align}
where $\Phi, \hat{\Phi}$ are the \emph{Schr\"odinger factors} that satisfy density factorization: $p_{t}(\bx) = \Phi_{t}(\bx) \hat{\Phi}_{t}(\bx)$\js{just a nitpick, but $\propto$ can also work, as score does not depend on normalization constants}. The vector-valued quantities $\bz_{t} = g(t) \nabla_{\bx} \log \Phi_{t}(\bx), \hat{\bz}_{t} = g(t) \nabla_{\bx} \log \hat{\Phi}_{t}(\bx)$ fully characterize dynamics of the SBP, thus can be considered as the forward, backward ``policies'', analogous to policy-based methods described in \cite{schulman2015trust,pereira2019neural}. To draw a link between SBPs and SGMs, the data log-likelihood objective for SBPs is computed and shown to be equal to that of SGMs with special choices of $\bz_{t}, \hat{\bz}_{t}$ (derivation details in \cite{chen2021likelihood}). Importantly, likelihood equality occurs with the following policies:
\begin{align}
    ( \bz_{t}, \hat{\bz}_{t} ) = ( 0, g(t) \, \nabla_{\bx} \log p_{t}(\bx) )
    \label{eq:policies}
\end{align}
When the marginal $p_1$ at time $t=1$ is equal to the prior distribution, it is known that such $( \bz_{t}, \hat{\bz}_{t} )$ are achieved. Since in SGMs, the end marginal $p_1$ is indeed the standard Gaussian prior, their log-likelihood is equivalent to that of SBPs. This suggests that SGMs are a special case of SBPs with degenerate forward policy $\bz_{t}$ and a multiple of the score function as its backward $\hat{\bz}_{t}$.
\se{couldn't quite follow this paragraph?}

\paragraph{Probability Flow ODE} In a similar vein to the SGM SDEs, a deterministic PF ODE can be derived for SBPs with identical marginal densities across $t \in [0, 1]$. The following PF ODE specifies the probability flow of the optimal processes of the SBP defined in \Cref{eq:sbp_sde,eq:sbp_sde} \citep{chen2021likelihood}:
\begin{align}
    \dd{\bx} = \left[ \bff(\bx, t) + g(t) \, \bz - \frac{1}{2} g(t) (\bz + \hat{\bz}) \right] \dd{t} \label{eq:pfode_sbp}
\end{align}
where $\bz$ depends on $\bx$. We shall show that the PF ODEs for SGMs and SBPs are equivalent. Thus, flowing through the PF ODEs in \modelshort{} is equivalent to flowing through special Schr\"odinger Bridges, with one of the marginals being Gaussian.
\se{didn't get the takeaway/message of this paragraph}

\section{\model{}}
\label{sec:method}

\begin{algorithm}[tb]
   \caption{High-level Pseudo-code for \modelshort{}}
   \label{alg:ddib}
\begin{algorithmic}
   \STATE {\bfseries Input:} data sample from source domain $\bx^{(s)} \sim p_{s}(\bx)$, source model $v^{(s)}_\theta$, target model $v^{(t)}_\theta$.
   \STATE {\bfseries Output: } $\bx^{(t)}$, the  result in the target domain.

   \STATE $\bx^{(l)} = \mathrm{ODESolve}(\bx^{(s)}; v^{(s)}_\theta, 0, 1)$  \quad // obtain latent code from source domain data
   \STATE $\bx^{(t)} = \mathrm{ODESolve}(\bx^{(l)}; v^{(t)}_\theta, 1, 0)$  \quad // obtain target domain data from latent code

   \STATE {\bfseries return}  $\bx^{(t)}$
\end{algorithmic}
\end{algorithm}

\modelshort{} leverage the connections between SGMs and SBPs to perform image-to-image translation, with two diffusion models trained separately on the two domains. 
\modelshort{} contain two steps, described in \Cref{alg:ddib} and illustrated in \Cref{fig:figure_1}. At the core of the algorithm is the ODE solver $\mathrm{ODESolve}$ from \Cref{eq:odesolve}. Given a source model represented as a vector field, \textit{i.e.}, $v^{(s)}_{\theta}$ defined from \Cref{eq:pfode_sgm},
\modelshort{} first apply $\mathrm{ODESolve}$ in the source domain to obtain the encoding $\bx^{(s)}$ of the image at the end time $t = 1$; we refer to this as the \textit{latent code} (associated with the diffusion model for the domain). Then, the source latent code is fed as the initial condition (target latent code at $t = 1$) to $\mathrm{ODESolve}$ with the target model $v^{(t)}_{\theta}$ to obtain the target image $\bx^{(t)}$. As discussed earlier, we implement $\mathrm{ODESolve}$ with DDIMs~\citep{song2020denoising}, which are known to have reasonably small discretization errors. While recent developments in higher order ODE solvers~\citep{zhang2022fast,lu2022dpm,karras2022elucidating} that generalize DDIMs can also be used here, we leave this investigation to future work.

Despite the simplicity of the method, \modelshort{} have several advantages over prior methods, which we discuss below.

\paragraph{Exact Cycle Consistency}
A desirable feature of image translation algorithms is the \emph{cycle consistency} property: transforming a data point from the source domain to the target domain, and then back to source, will recover the original data point in the source domain. The following proposition validates the cycle consistency of \modelshort{}.

\begin{proposition}[\modelshort{} Enforce Exact Cycle Consistency] Given a sample from source domain $\bx^{(s)}$, a source diffusion model $v^{(s)}_\theta$, and a target model $v^{(t)}_\theta$, define:
\begin{align}
    \bx^{(l)} = \mathrm{ODESolve}(\bx^{(s)}; v^{(s)}_\theta, 0, 1); &\quad \bx^{(t)} = \mathrm{ODESolve}(\bx^{(l)}; v^{(t)}_\theta, 1, 0);\\
    \bx'^{(l)} = \mathrm{ODESolve}(\bx^{(t)}; v^{(t)}_\theta, 0, 1); &\quad \bx'^{(s)} = \mathrm{ODESolve}(\bx'^{(l)}; v^{(s)}_\theta, 1, 0)
\end{align}
Assume zero discretization error. Then, $\bx^{(s)} = \bx'^{(s)}$.
\end{proposition}
\se{why is there a trained assumption in the statement. i think this is true even for an untrained model? basically, no assumptions on how well these models are trained}
As PF ODEs are used, the cycle consistency property is guaranteed. In practice, even with discretization error, DDIBs incur almost negligible cycle inconsistency (\Cref{sec:2d_experiments}). In contrast, GAN-based methods are not guaranteed the cycle consistency property by default, and have to incorporate additional training terms to optimize for cycle consistency over two domains.
\se{is this true? i thought you can enforce with unpaired data}

\paragraph{Data Privacy in Both Domains} In the \modelshort{} translation process, only the source and target diffusion models are required, whose training processes do not depend on knowledge of the domain pair \emph{a priori}. In fact, this process can even be performed in a privacy sensitive manner (graphic illustration in \Cref{appendix:privacy_sensitive}). Let Alice and Bob be the data owners of the source and target domains, respectively. Suppose Alice intends to translate images to the target domain. However, Alice does not want to share the data with Bob (and vice versa, Bob does not want to release their data either). Then, Alice can simply train a diffusion model with the source data, encode the data to the latent space, transmit the latent codes to Bob, and next ask Bob to run their trained diffusion model and send the results back. In this procedure, only the latent code and the target results are transmitted between the two data vendors, and both parties have naturally ensured that their data are not directly revealed.

\paragraph{DDIBs are Two Concatenated Schr\"odinger Bridges} \modelshort{} link the source data distribution to the latent space, and then to the target distribution. What is the nature of such connections between distributions? We offer an answer from an optimal transport perspective: these connections are special \emph{Schr\"odinger Bridges} between distributions.
This, in turn, explicates the name of our method: \modellower{} are based on denoising \emph{diffusion implicit} models \citep{song2020denoising}, and consist of \emph{two} separate Schr\"odinger \emph{Bridges} that connect the data and latent distributions.\lingxiao{What is the benefit of having this viewpoint? Does using OT imply better domain translation? I guess this is discussed below in particular in ``exact cycle consistency".} Specifically, as considered earlier, when conditions\lingxiao{what conditions? cite or expand} about the policies $\bz_{t}, \hat{\bz}_{t}$ in \Cref{eq:policies} and the density $p_{1}(\bx)$ being a Gaussian prior are met, the data likelihoods (at $t=0$) for SGMs and SBPs are identical\lingxiao{This is likelihood at time 0?}. Indeed, these conditions are fulfilled in SGMs and particularly in DDIMs. This verifies SGMs as special linear or degenerate SBPs. Forward and reverse solving the PF ODE for SGMs, as done in \modelshort{}, is equivalent to flowing through the optimal processes of particular SBPs:
\begin{proposition}[PF ODE Equivalence\footnote{Proof in \Cref{appendix:proof}.}] \Cref{eq:pfode_sgm} is equivalent to \Cref{eq:pfode_sbp} with forward, backward policies $(\bz_{t}, \hat{\bz}_{t}) = (0, g \nabla_{\bx} \log p_{t}(\bx))$ as attained in SGMs\js{that take the PF-ODE form} and particularly in DDIMs\lingxiao{(2) does not have $z_t, \hat z_t$, so this really means when plugging in the policies for (5) they agree?}.
\label{prop:ode_equiv}
\end{proposition}
\se{don't quite understand the meaning of this. can you rephrase? specifially the equations referenced have nothing to do with domain translation (it's for single domain case), so how can this proposition even say something about translation?}

Thus, \modelshort{} are intrinsically entropy-regularized optimal transport: they are Schr\"odinger Bridges between the source and the latent, and between the latent and the target distributions. The translation process can then be recognized as traversing through two concatenated Schr\"odinger Bridges, one forward and one reversed. The mapping is unique and minimizes a (regularized) optimal transport objective, which probably elucidates the superior performance of \modelshort{}. In contrast, if we train the source and target models separately with normalizing flow models that are not inborn with such a connection, there are many viable invertible mappings, and the resulting image translation algorithm may not necessarily have good performance. This is probably the reason why AlignFlow~\citep{grover2020alignflow} still has to incorporate an adversarial loss even when cycle-consistency is guaranteed.

\section{Experiments}

We present a series of experiments to demonstrate the effectiveness of \modelshort{}. First, we describe synthetic experiments on two-dimensional datasets, to corroborate \modelshort{}' cycle-consistent and optimal transport properties. Next, we validate \modelshort{} on a variety of image translation tasks, including color transfer, paired translation, and conditional ImageNet translation.
\footnote{\textbf{Project}: \url{https://suxuann.github.io/ddib/}}\footnote{\textbf{Code}: \url{https://github.com/suxuann/ddib/}}


\subsection{2D Synthetic Experiments}
\label{sec:2d_experiments}

\begin{figure}[!ht]
\vspace{-.2cm}
\subfloat[Smooth translation of synthetic datasets. (\textit{Left}) The source datasets: CR and CS. (\textit{Middle}) \modelshort{}' latent code representation. (\textit{Right}) Results of translation to the target domains. \label{fig:2d_translation}]{
  \includegraphics[width=.48\textwidth]{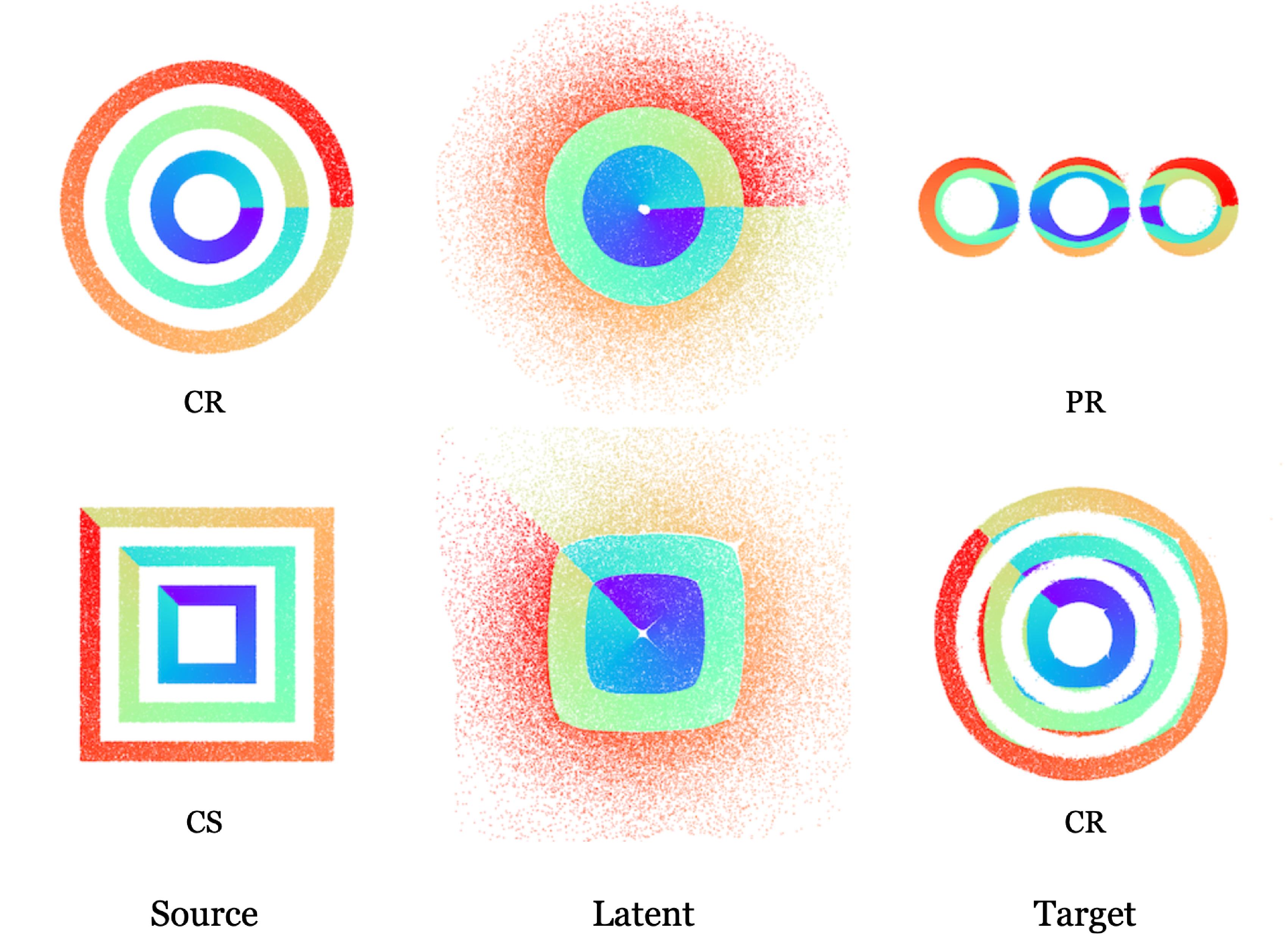}
}
\hfill
\subfloat[\textbf{Cycle consistency}: After translating the Moons dataset to Checkerboards and then back to Moons, \modelshort{} restore almost the exact same points as the original ones.\label{fig:2d_cycle}]{
  \includegraphics[width=.48\textwidth]{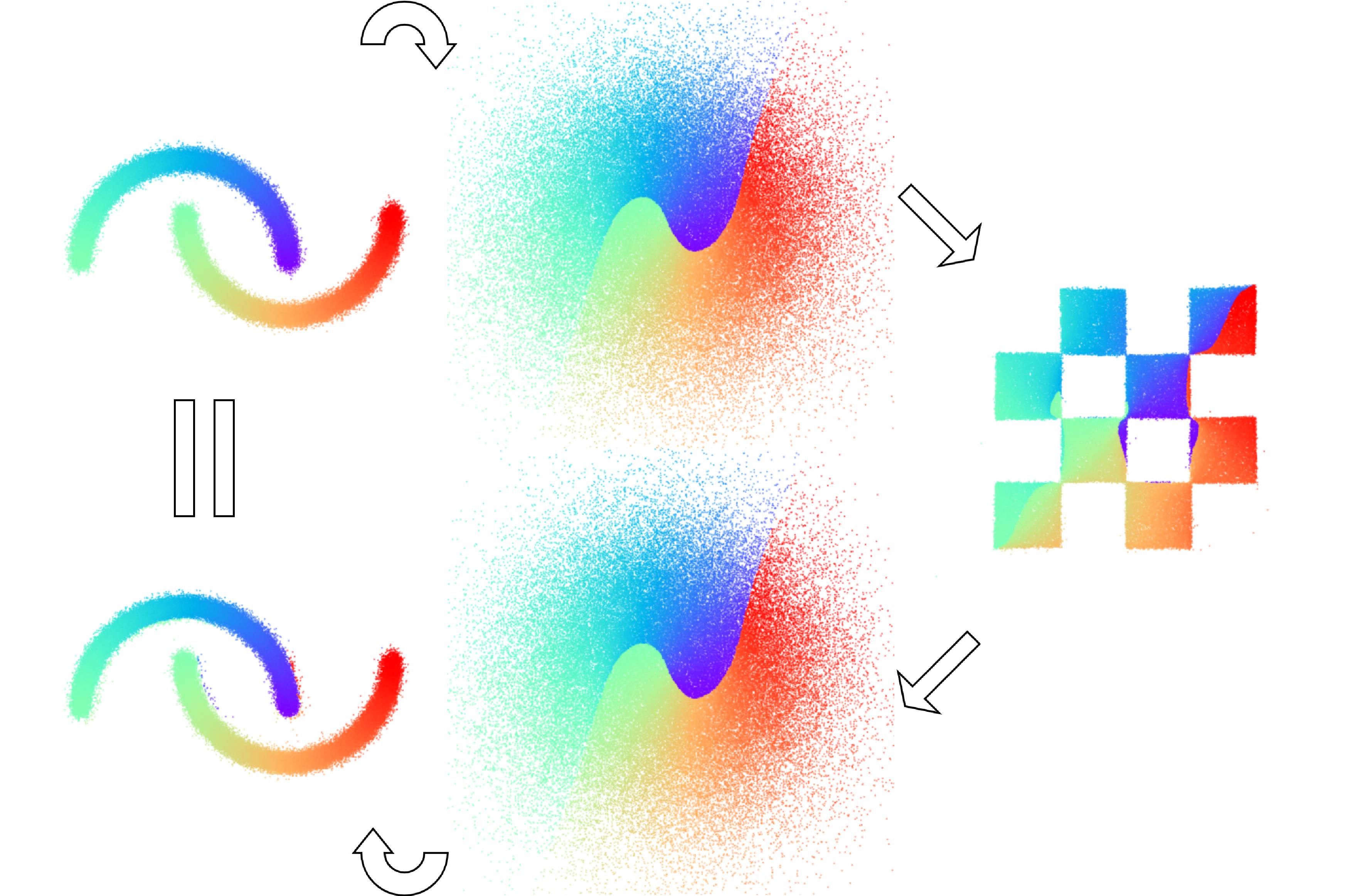}
}
\caption{Smoothness and cycle consistency of \modelshort{}.}
\label{fig:dummy}
\vspace{-.4cm}
\end{figure}

We first perform domain translation on synthetic datasets drawn from complex two-dimensional distributions, with various shapes and configurations, in \Cref{fig:2d_translation}. In total, we consider six 2D datasets: Moons (M); Checkerboards (CB); Concentric Rings (CR); Concentric Squares (CS); Parallel Rings (PR); and Parallel Squares (PS). The datasets are all normalized to have zero mean, and identity covariance. We assign colors to points based on the point identities (\textit{i.e.}, if a point in the source domain is red, its corresponding point in the target domain is also colored red). Clearly, the transformation is \emph{smooth} between columns. For example, on the top-right corner, red points in the CR dataset are mapped to similar coordinates, both in the latent and in the target dimensions.

\begin{table}[t]
\caption{Cycle consistency of \modelshort{}. Experiment legend, PR $\circlearrow$ PS, means that we translate from PR to PS and then back. The numbers are the averaged L2 distances between the original points and their coordinates after cycle translation. Data points are standardized to have unit variance.}
\label{table:cycle_consistency}
\begin{center}
\begin{small}
\begin{sc}
\begin{tabular}{ccccc}
\toprule
PR $\circlearrow$ PS & PS $\circlearrow$ CS & CR $\circlearrow$ PR & CR $\circlearrow$ CS & M $\circlearrow$ CB \\
\midrule
0.0143 & 0.0065 & 0.0106 & 0.0078 & 0.0122\\
\bottomrule
\end{tabular}
\end{sc}
\end{small}
\end{center}
\vskip -0.1in
\end{table}


\paragraph{Cycle Consistency} \Cref{fig:2d_cycle} illustrates the cycle consistency property guaranteed by \modelshort{}. It concerns 2D datasets: Moons, and Checkerboards. Starting from the Moons dataset, \modelshort{} first obtain the latent codes and construct the Checkerboards points. Next, \modelshort{} do translations in the reverse direction, transforming the points back to the latent and the Moons space. After this round trip, points are approximately mapped to their original positions. A similar, smooth color topology is observed in this experiment. \Cref{table:cycle_consistency} reports quantitative evaluation results on cycle-consistent translation among multiple datasets. As the datasets are normalized to unit standard deviation, the reported values are negligibly small and endorse the cycle consistent property of \modelshort{}.\js{draw a conclusion: the data standard deviation is 1 so the error within the cycle is small (maybe cite some data on MSE for trained cycle consistency) and also note that MSE is L2-squared, so your actual MSE would be much smaller. also it might help to be consistent with L2 and MSE throughout the paper}


\subsection{Example-Guided Color Transfer}

\begin{figure*}[ht]
\begin{center}
\centerline{\includegraphics[width=\linewidth]{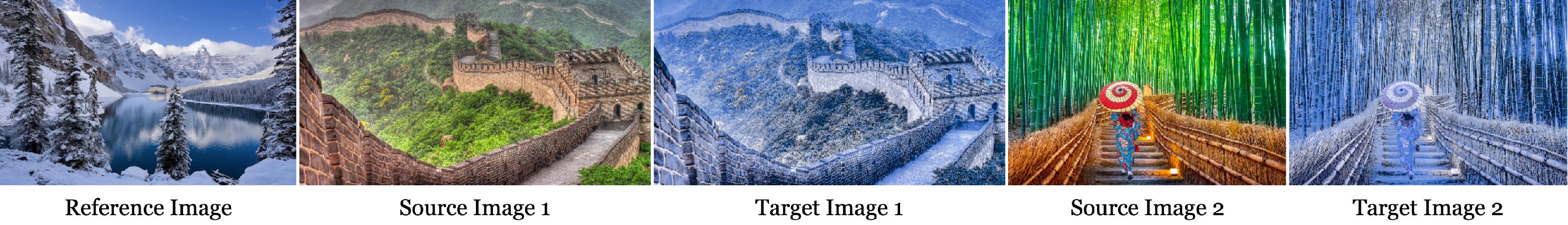}}
\caption{\textbf{Example-Guided Color Transfer}: Given the first image as the reference image, \modelshort{} modify the colors of two input images to similarly follow a snowy winter color palette.}
\label{fig:color_transfer}
\end{center}
\vskip -0.3in
\end{figure*}

\modelshort{} can be used on an interesting application: example-guided color transfer. This refers to the task of modifying the colors of an input image, conditioned on the color palette of a reference image. To use \modelshort{} for color transfer, we train one diffusion model per image, on its normalized RGB space. During translation, \modelshort{} obtain encodings of the original colors, and apply the diffusion model of the reference image to attain the desired color palette. \Cref{fig:color_transfer} visualizes our color experiments.

\paragraph{Comparison to Alternative OT Methods} As \modelshort{} are related to regularized OT, we compare the pixel-wise MSEs between color-transferred images generated by \modelshort{}, and images produced by alternate methods, in \cref{table:color_transfer}. We include four OT methods for comparison: Earth Mover's Distance; Sinkhorn distance \citep{cuturi2013sinkhorn}; linear and Gaussian mapping estimation \citep{perrot2016mapping}. Results of \modelshort{} are very close to those of OT methods. \Cref{appendix:color-transfer} details full color translation results.

\begin{table}[t]
\caption{Mean Squared Error (MSE) comparing color transfer results of \modelshort{} with common OT methods on two images. Each number represents the MSE between \modelshort{} and the corresponding OT method. MSE is computed pixel-wise after normalizing images to $[-1, 1]$.}
\label{table:color_transfer}
\begin{center}
\begin{small}
\begin{sc}
\begin{tabular}{l|cccc}
\toprule
Image & EMD & Sinkhorn & Linear & Gaussian \\
\midrule
Target 1 & 0.0337 & 0.0281 & 0.0352 & 0.0370 \\
Target 2 & 0.0293 & 0.0326 & 0.0500 & 0.0751 \\
\bottomrule
\end{tabular}
\end{sc}
\end{small}
\end{center}
\vskip -0.1in
\end{table}

\subsection{Quantitative Translation Evaluation}
Quantitatively, we demonstrate that \modelshort{} deliver competitive results on paired domain tests. 
Such numerical evaluation is despite that \modelshort{} are formulated with a weaker setting: diffusion models are trained independently, on separate datasets. In comparison, methods such as CycleGAN and AlignFlow assume access to both datasets during training and jointly optimize for the translation loss.

\begin{table}[t]
\caption{MSE comparing \modelshort{} and baselines on paired test sets. MSE is computed pixel-wise after normalizing images to $[-1, 1]$.}
\label{table:mse}
\begin{center}
\begin{small}
\begin{sc}
\begin{tabular}{l|ccc|c|cccr}
\toprule
Dataset & Model & \hspace{-0.25cm} A $\rightarrow$ B & \hspace{-0.25cm} B $\rightarrow$ A & Dataset & Model & \hspace{-0.25cm} A $\rightarrow$ B & \hspace{-0.25cm} B $\rightarrow$ A \\
\midrule
\multirow{3}{*}{Facades}    & CycleGAN  & 0.7129 & 0.3286 & \multirow{3}{*}{Maps}       & CycleGAN  & 0.0245 & 0.0953 \\
           & AlignFlow & 0.5801 & \textbf{0.2512} &            & AlignFlow & 0.0209 & \textbf{0.0897} \\
           & \modelshort{}      & \textbf{0.5312} & 0.3946 & & \modelshort{}      & \textbf{0.0194} & 0.1302\\
\bottomrule
\end{tabular}
\end{sc}
\end{small}
\end{center}
\vskip -0.2in
\end{table}

\paragraph{Paired Domain Translation} As in similar works, we evaluate \modelshort{} on benchmark paired datasets \citep{zhu2017unpaired}: Facades and Maps. Both are image segmentation tasks. In the pairs of datasets, one dataset contains real photos taken via a camera or a satellite; while the other comprises the corresponding segmentation images. These datasets provide one-to-one image alignment, which allows quantitative evaluation through a distance metric such as mean-squared error (MSE) between generated samples and the corresponding ground truth. To facilitate the workings of \modelshort{}, we additionally employ a color conversion heuristic motivated by optimal transport on image colors (\Cref{appendix:color-conversion}). \Cref{table:mse} reports the evaluation results. Surprisingly, \modelshort{} are able to produce segmentation images that surpass alternative methods in MSE terms; while reverse translations also achieve decent performance.

\subsection{Class-Conditional ImageNet Translation}

\begin{figure}[!ht]
\subfloat[\textbf{Conditional ImageNet Translation}: Selected translation samples from various ImageNet classes such as 7: Cock, 94: Hummingbird, 162: Beagle, and 282: Tiger Cat.\label{fig:imagenet_1}]{
  \includegraphics[width=.49\textwidth]{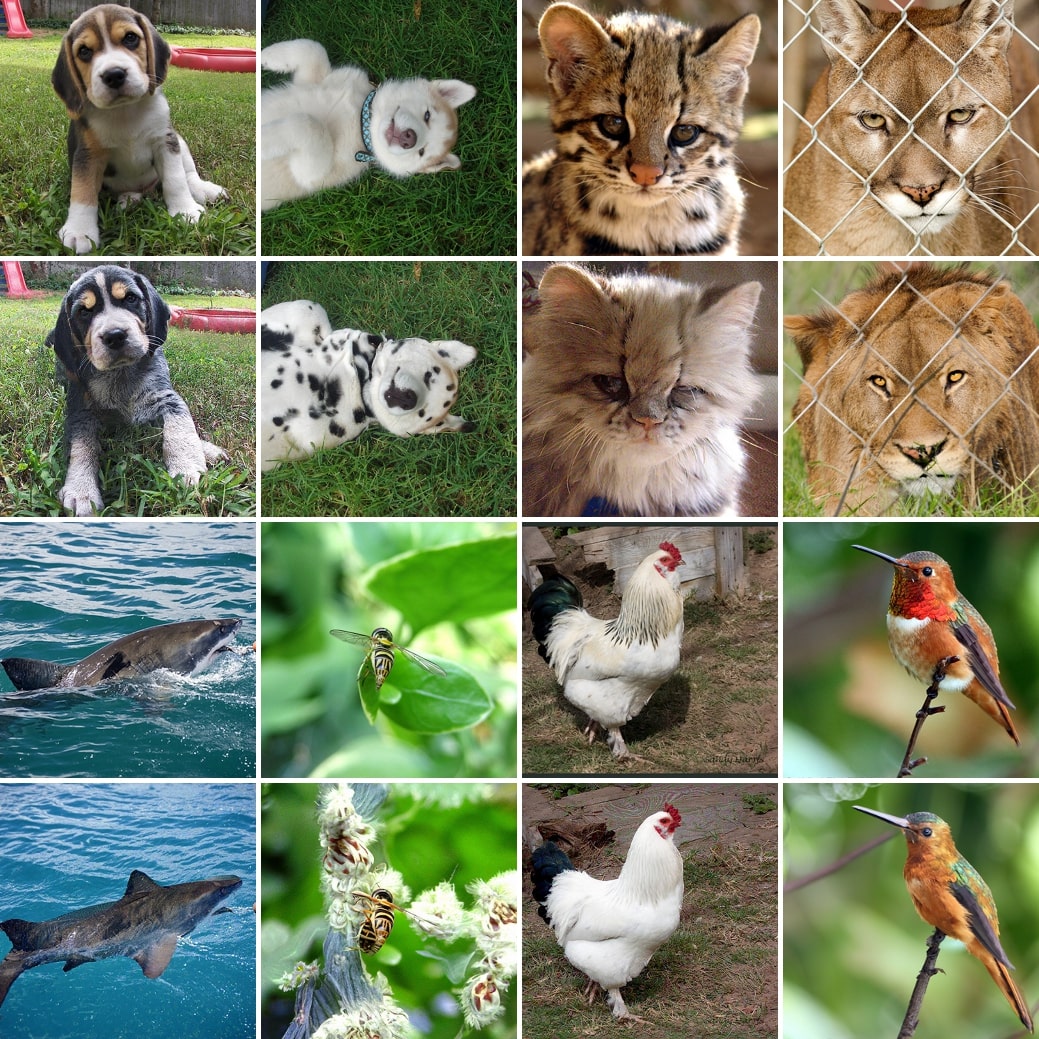}
}
\hfill
\subfloat[\textbf{Multi-domain translation}: Given the center, source image from class label 291, \modelshort{} translate it to other animal species, entirely using only a pretrained conditional diffusion model.\label{fig:multidomain}]{
  \includegraphics[width=.47\textwidth]{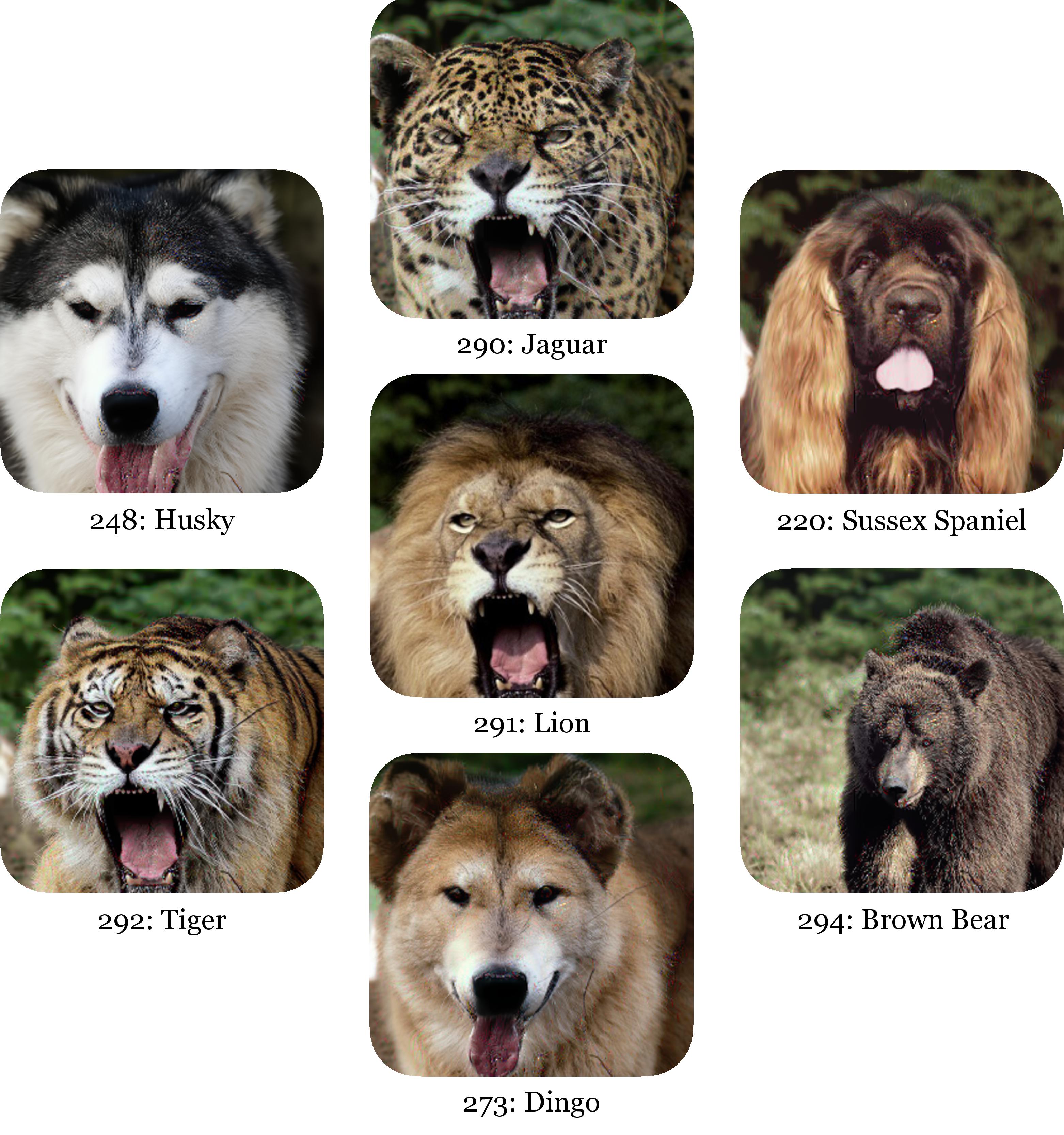}
}
\label{fig:imagenet}
\vspace{-.2cm}
\end{figure}

In this experiment, we apply \modelshort{} to translation among ImageNet classes. To this end, we leverage the pretrained diffusion models from \cite{dhariwal2021diffusion}. The authors optimized performance of diffusion models, and end up with a ``UNet'' \citep{ho2020denoising} architecture with particular width, attention and residual configurations. The models are learned on $1,000$ ImageNet classes, each with around $1,000$ training images, and at a variety of resolutions. Our experiments use the model with resolution $256 \times 256$. Moreover, these models incorporate a technique known as \emph{classifier guidance} \citep{dhariwal2021diffusion}, that leverage classifier gradients to steer the sampling process towards arbitrary class labels during image generation. The learned models combined with classifier guidance can be effectively considered as $1,000$ different models. \Cref{fig:imagenet_1} exhibits select translation samples, where the source images are from ImageNet validation sets. \modelshort{} are able to create faithful target images that maintain much of the original content such as animal poses, complexions and emotions, while accounting for differences in animal species.

\paragraph{Multi-Domain Translation} Given conditional models on the individual domains, \modelshort{} can be applied to translate between arbitrary pairs of source-target domains, while requiring no additional fine-tuning or adaptation. \Cref{fig:multidomain} displays results of translating a common image of a roaring lion (with class label 291), to various other ImageNet classes. Interestingly, some animals roar, while others stick their tongues out. \modelshort{} successfully internalize characteristics of distinct animal species, and produce closest animal postures in OT distances to the original shouting lion.

\begin{figure*}[ht]
\begin{center}
\centerline{\includegraphics[width=\linewidth]{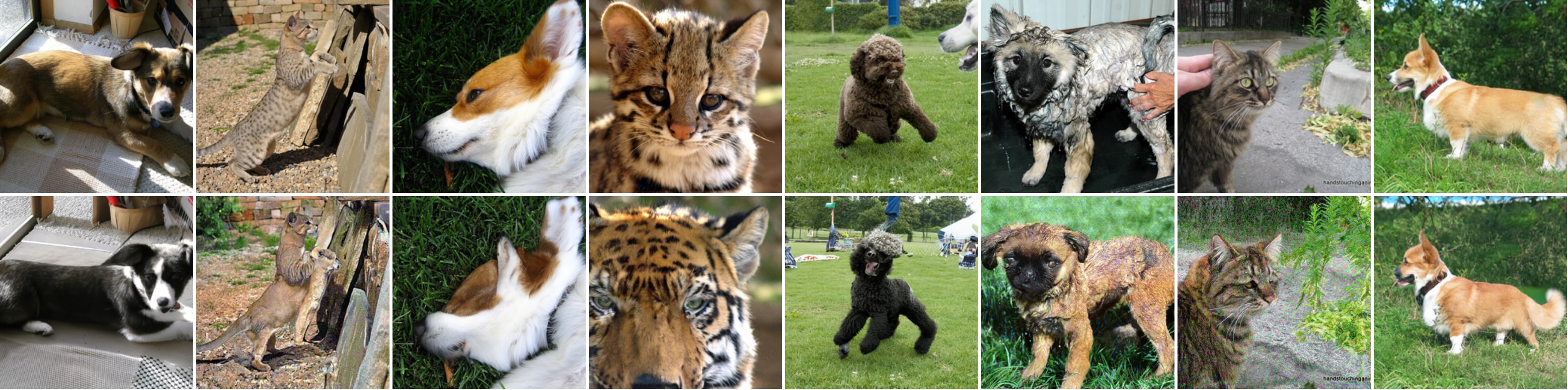}}
\caption{Translation among ImageNet classes.}
\label{fig:additional_image_net}
\end{center}
\vskip -0.3in
\end{figure*}

\section{Related Works}

\paragraph{Score-based Diffusion Models} Originating in thermodynamics \citep{sohl2015deep}, diffusion models reverse the dynamics of a noising process to create data samples. The reversal process is understood to implicitly compute scores of the data density at various noise scales, which reveals connections to score-based methods \citep{song2019generative,nichol2021improved,meng2021estimating}. Diffusion models are applicable to multiple modalities: 3D shapes \citep{zhou20213d}, point cloud \citep{luo2021diffusion}, discrete domains \citep{meng2022concrete} and function spaces \citep{lim2023score}. They excel in tasks ranging from image editing and composition \citep{meng2021sdedit}, density estimation \citep{kingma2021variational}, to image restoration \citep{kawar2022denoising}. Seminal works are \emph{denoising diffusion probabilistic models} (DDPMs, \cite{ho2020denoising}), which parameterized the ELBO objective with Gaussians and, for the first time, synthesized high-quality images with diffusion models; ILVR \citep{choi2021ilvr}, which invented a novel conditional method to direct DDPM generation towards reference images; and \emph{denoising diffusion implicit models} (DDIMs, \cite{song2020denoising}), which accelerated DDPM inference via non-Markovian processes. DDIMs can be treated as a first-order numerical solver of a probabilistic ODE, which we use heavily in \modelshort{}.


\paragraph{Diffusion Models for Image Translation} While GANs \citep{goodfellow2014generative,zhu2017unpaired,zhao2020unpaired} have been widely adopted in image translation tasks, recent works increasingly leverage diffusion models. For instance, Palette \citep{saharia2021palette} applies a conditional diffusion model to colorization, inpainting, and restoration. DiffuseIT \citep{kwon2022diffusion} utilizes disentangled style and content representation, to perform text- and image-guided style transfer. Lastly, UNIT-DDPM \citep{sasaki2021unit} proposes a novel coupling between domain pairs and trains joint DDPMs for translation. Unlike their joint training, \modelshort{} apply separate, pretrained diffusion models and leverage geometry of the shared space for translation.

\paragraph{Optimal Transport for Translation and Generative Modeling} As it pursues cost-optimal plans to connect image distributions, OT naturally finds applications in image translation. For example, \cite{korotin2022neural} capitalizes on the approximation powers of neural networks to compute OT plans between image distributions and perform unpaired translation. By contrast, the entropy-regularized OT variant, Schr\"odinger Bridges (\Cref{sec:preliminaries}), are also commonly used to derive generative models. For instance, \cite{de2021diffusion} and \cite{vargas2021solving} concurrently proposed new numerical procedures that approximate the Iterative Proportional Fitting scheme, to solve SBPs for image generation. \cite{wang2021deep} presents a new generative method via entropic interpolation with an SBP. \cite{chen2021likelihood} discovers equivalence between the likelihood objectives of SBP and score-based models, which lays the theoretical foundations behind \modelshort{}. Their sequel \citep{liu20232} then directly learns the Schr\"odinger Bridges between image distributions, for applications in image-to-image tasks such as restoration. While \modelshort{} were not initially designed to mimic Schr\"odinger Bridges, our analysis reveals their true characterization as solutions to degenerate SBPs.

\section{Conclusions}
We present \model{} (\modelshort{}), a new, simplistic image translation method that stems from latest progresses in score-based diffusion models, and is theoretically grounded as Schr\"odinger Bridges in the image space. \modelshort{} solve two key problems. First, \modelshort{} avoid optimization on a coupled loss specific to the given domain pair only. Second, \modelshort{} better safeguard dataset privacy as they no longer require presence of both datasets during training. Powerful pretrained diffusion models are then integrated into our \modelshort{} framework, to perform a comprehensive series of experiments that prove \modelshort{}' practical values in domain translation. Our method is limited in its application to color transfer, as one model is required for each image, which demands significant compute for mass experiments. Rooted in optimal transport, \modelshort{} translation mimics the mass-moving process which may be problematic at times (\Cref{appendix:limitations}). Future work may remedy these issues, or extend \modelshort{} to applications with different dimensions in the source and target domains. As flowing through the concatenated ODEs is time-consuming, improving the translation speed is also a promising direction.

\section*{Acknowledgements}

We thank Lingxiao Li and Chris Cundy for insightful discussions about the optimal transport properties of \modelshort{}. We also thank the anonymous reviewers for their constructive comments and feedback. This research was supported by NSF (\#1651565), ARO (W911NF-21-1-0125), ONR (N00014-23-1-2159), CZ Biohub, and Stanford HAI.

\bibliography{iclr2023_conference}
\bibliographystyle{iclr2023_conference}

\appendix
\newpage

\appendix
\onecolumn

\section{Illustration: Privacy-Sensitive Translation}
\label{appendix:privacy_sensitive}

\begin{wrapfigure}{l}{0.53\textwidth}
  \vspace{-.5cm}
  \begin{center}
    \includegraphics[width=.53\textwidth]{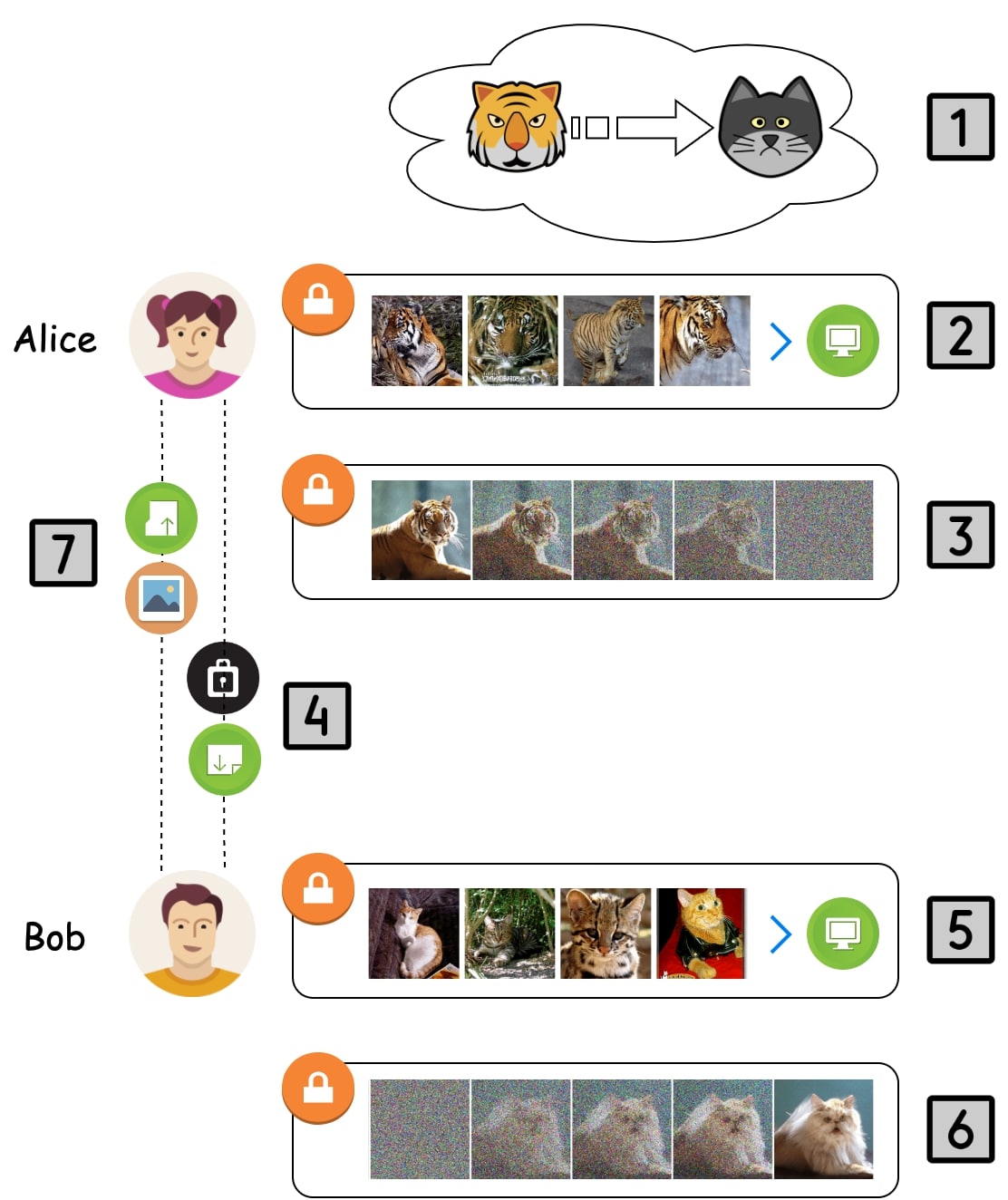}
  \end{center}
  \caption{}
  \label{fig:privacy_sensitive}
\end{wrapfigure}

Alice is the owner of the source (tiger) domain, and Bob is the owner of the target (cat) domain. Alice intends to translate tiger images to cat images, but in a privacy-sensitive manner without releasing the source dataset. Bob does not wish to make the cat dataset public, either.

\cref{fig:privacy_sensitive} illustrates the process of privacy-sensitive domain translation. The process contains the following steps, with indexes in the figure.
\begin{enumerate}
    \item Alice intends to translate tiger images to cat images.
    \item Alice trains a diffusion model with the source tiger images.
    \item Alice uses the pretrained, tiger diffusion model to convert a source tiger image to its latent code.
    \item Alice sends the latent code to Bob.
    \item Bob similarly trains a diffusion model on the cat domain.
    \item Bob uses the pretrained, cat \\diffusion model to convert the received latent code to a cat image.
    \item Bob then sends the translated image back to Alice.
\end{enumerate}
Clearly, during the translation process, only the latent code and the translated cat image are transmitted via the public channel, while both source and target datasets are private to the two parties. This is a significant advantage of \modelshort{} over alternate methods, as we enable strong privacy protection of the datasets.

\newpage
\section{Details of SGM Training and DDIM ODE Solver}
\label{appendix:ddim_details}

\revision{
\subsection{Training Score Networks}
While the description in \Cref{sec:preliminaries} is based on continuous SDEs, actual implementations of diffusion models often sample discrete time steps. Given samples from a data distribution $q(\bx_0)$, diffusion models attempt to learn a model distribution $p_{\theta}(\bx_0)$ that approximates $q(\bx_0)$, and is easy to sample from. Specifically, diffusion probabilistic models are latent variable models of the form
\begin{align*}
    p_{\theta}(\bx_0) = \int p_{\theta}(\bx_{0:T}) \dd{\bx_{1:T}}, \text{ where } p_{\theta}(\bx_{0:T}) = p_{\theta}(\bx_{T}) \prod_{t=1}^{T} p_{\theta}^{(t)}(\bx_{t-1} | \bx_{t})
\end{align*}
where $\bx_1, \cdots, \bx_T$ are latent variables in the same sample space as $\bx_{0}$. The parameters $\theta$ are trained to approximate the data distribution $q(\bx_0)$, by maximizing a variational lower bound:
\begin{align*}
    \max_{\theta} \mathbb{E}_{q(\bx_0)} [ \log p_{\theta}(\bx_0) ] \leq \max_{\theta} \mathbb{E}_{q(\bx_0, \bx_1, \cdots, \bx_T)} [ \log p_{\theta}(\bx_{0:T}) - \log q(\bx_{1:T} | \bx_0) ] 
\end{align*}
where $q(\bx_{1:T} | \bx_0)$ is some inference distribution over the latent variables. It is known that when the conditional distributions are modeled as Gaussians with trainable mean functions and fixed variances, the above objective can be simplified to:
\begin{align*}
    L( \epsilon_{\theta} ) := \sum_{t=1}^{T} \mathbb{E}_{\bx_0 \sim q(\bx_0), \epsilon_t \sim \mathcal{N}(\bZero, \bI) } \left[ \norm{ \epsilon_{\theta}^{(t)} ( \sqrt{\alpha_t} \bx_0 + \sqrt{1 - \alpha_t} \epsilon_t ) - \epsilon_{t} }^{2}_{2} \right]
\end{align*}
The resulting noise prediction functions $\epsilon_{\theta}^{(t)}$, are equivalent to the score networks $\bs_{t,\theta}$ mentioned in \cref{sec:preliminaries} due to Tweedie's formula~\citep{stein1981estimation,efron2011tweedie}. For details, we refer the reader to \cite{ho2020denoising,song2020denoising}.

}

\subsection{DDIM ODE Solver}
With a trained noise prediction model $\epsilon^{(t)}_{\theta}(\bx)$, the DDIM iterate between adjacent variables $\bx_{t - \Delta t}$ and $\bx_{t}$, considered in \cite{song2020denoising}, assumes the following form:
\begin{align*}
    \frac{ \bx_{t - \Delta t} }{ \sqrt{ \alpha_{ t - \Delta t} } } = \frac{ \bx_{t} }{ \sqrt{ \alpha_{t} } } + \left( \sqrt{ \frac{ 1 - \alpha_{ t - \Delta t} }{ \alpha_{ t - \Delta t} } } - \sqrt{ \frac{ 1 - \alpha_{ t } }{ \alpha_{ t } } } \right) \epsilon^{(t)}_{\theta}(\bx_{t})
\end{align*}
In our experiments, we implement the above equation between adjacent diffusion steps. The equation is deterministic, and can be considered as a Euler method over the following ODE:
\begin{align}
    \dd{ \bar{\bx}(t) } = \epsilon^{(t)}_{\theta} \left( \frac{ \bar{\bx}(t) }{ \sqrt{\sigma^{2} + 1 }} \right) \dd{ \sigma(t) }
    \label{eq:ddim_ode}
\end{align}
where we adopt the reparameterization:
\begin{align*}
    \sigma(t) = \sqrt{ \frac{ 1 - \alpha(t) }{ \alpha(t) } }, \quad \bar{\bx}(t) = \frac{ \bx(t) }{ \sqrt{ \alpha(t) } }
\end{align*}
Importantly, the ODE in \Cref{eq:ddim_ode} with the optimal model $\epsilon^{(t)}_{\theta}(\bx)$, has an equivalent probability flow ODE corresponding to the ``Variance-Exploding'' SDE in \cite{song2020score}.

\newpage
\section{Limitations of Optimal Transport-Based Translation}
\label{appendix:limitations}

\modelshort{} contain deterministic bridges between distributions, and are a form of entropy-regularized optimal transport. The learned diffusion models can be effectively considered as a digest or summary of the datasets. While doing translation, they attempt to create images in the target domain, that are closest in optimal transport distances to the source images. Such OT-based process is both an advantage and a limitation of our method.

In ImageNet translation, when the source and target datasets are similar, DDIBs are generally able to identify correct animal postures. For example, we have shouting lions and tigers, because these animals have similar behaviors that are observed in the datasets and then internalized by DDIBs. However, in datasets that are less similar (\textit{e.g.} birds and dogs), \modelshort{} sometimes fail to produce translation results that retain the postures precisely. We encountered significantly less such cases in AFHQ translation, since the dataset is more standardized and homogeneous.

\begin{figure*}[ht]
\begin{center}
\centerline{\includegraphics[width=\linewidth]{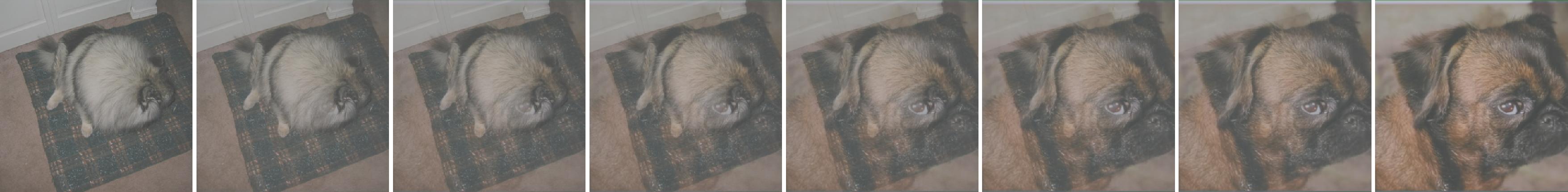}}
\vskip 0.03in
\centerline{\includegraphics[width=\linewidth]{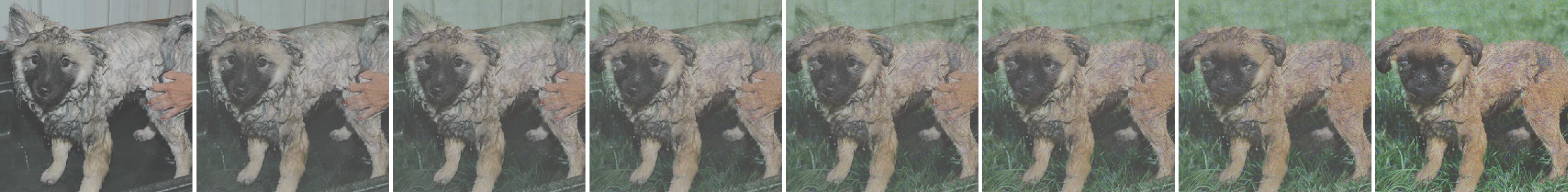}}
\vskip 0.03in
\centerline{\includegraphics[width=\linewidth]{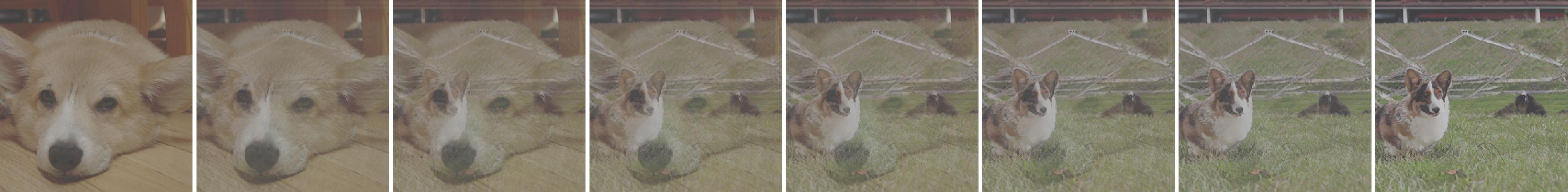}}
\vskip 0.03in
\centerline{\includegraphics[width=\linewidth]{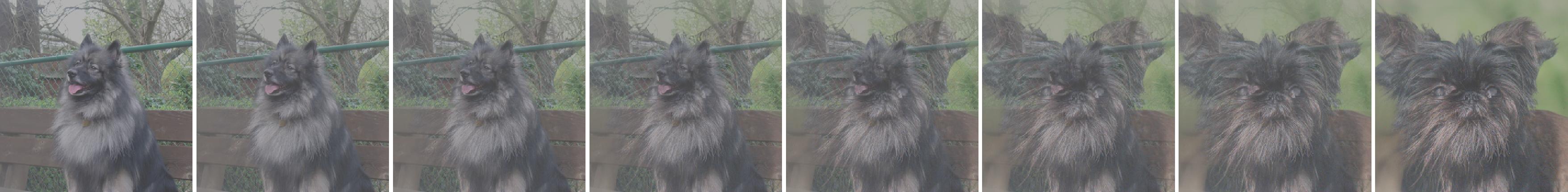}}
\vskip 0.03in
\caption{Optimal transport translation processes in \modelshort{}. \textit{(Leftmost)} Source images. \textit{(Rightmost)} Translated images.}
\label{fig:ot_illustration}
\end{center}
\end{figure*}

\Cref{fig:ot_illustration} illustrates the optimal transport mappings among images as well as some failure cases. Clearly, the translation processes flowing from left to right minimize the Euclidean transportation distances between images. Some of these translated samples may be classified ``failure cases'' in actual user studies. Such are considered both a feature and a limitation of \modelshort{}.

\newpage
\section{Proof of \Cref{prop:ode_equiv}}
\label{appendix:proof}

\begin{proof}
The proof proceeds by substituting the values of $(\bz_{t}, \hat{\bz}_{t}) = (0, g(t) \nabla_{\bx} \log p_{t}(\bx))$ into \Cref{eq:pfode_sbp},
\begin{align}
    \dd{\bx} &= \left[ \bff(\bx, t) + g(t) \, \bz - \frac{1}{2} g(t) (\bz + \hat{\bz}) \right] \dd{t}\\
    &= \left[ \bff(\bx, t) - \frac{1}{2} g(t)^{2} \nabla_{\bx} \log p_{t}(\bx) \right] \dd{t}
\end{align}
This is exactly \Cref{eq:pfode_sgm}.
\end{proof}

\newpage
\section{Additional Experimental Details}
\label{appendix:experiments}

\subsection{Optimal Transport in Paired Datasets}
\label{appendix:color-conversion}

\begin{figure}[ht]
\vskip 0.2in
\begin{center}
\centerline{\includegraphics[width=0.6\columnwidth]{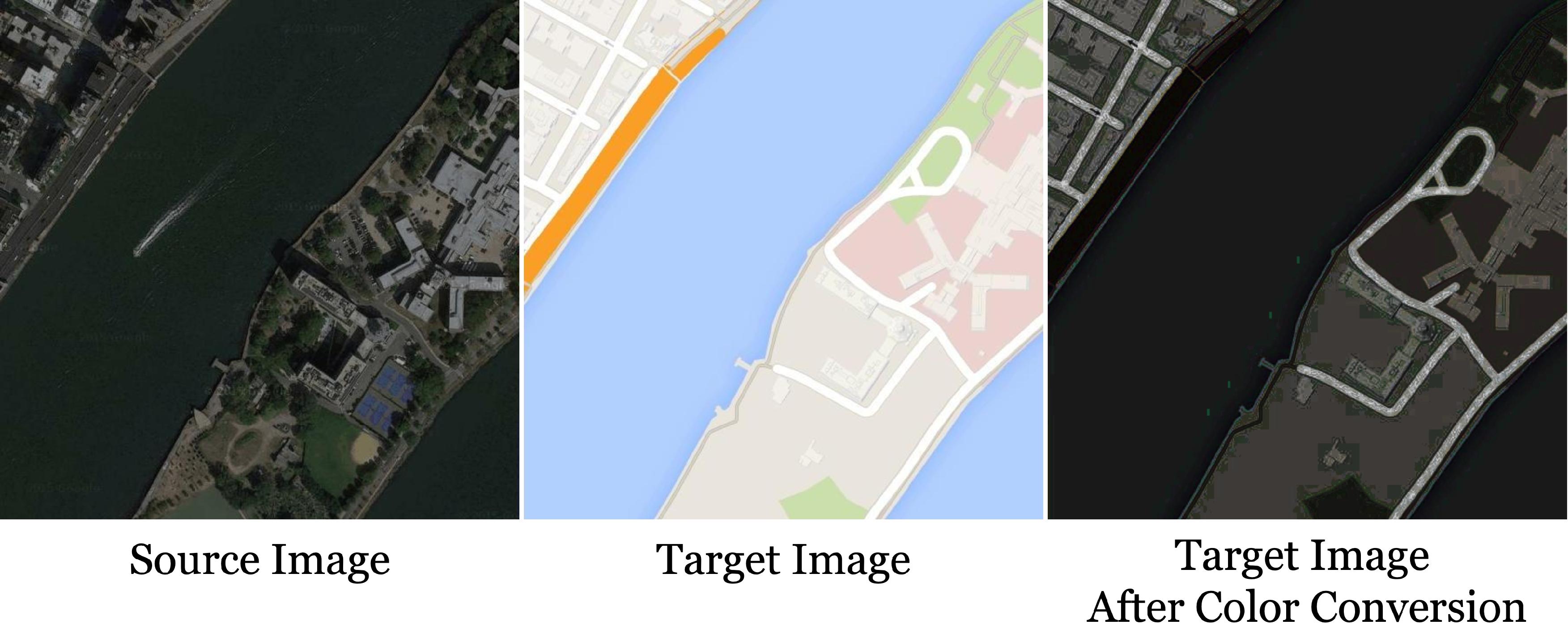}}
\caption{\textbf{Color Conversion}. In the paired translation tasks, we are given the real and segmentation images. Before training the diffusion models, we first transform the segmentation images to a color palette that is closer to the real images. While evaluating MSEs, we convert the images back to the original colors.}
\label{fig:color_conversion}
\end{center}
\vskip -0.2in
\end{figure}

\paragraph{Color Conversion} In \Cref{fig:color_conversion}, a simple examination of the original and segmentation images reveals significant differences in color configurations. In the Maps dataset, while the real, satellite images are composed of dark colors, the segmentation images are light-toned. The same observation applies to other datasets. The shark contrasts in colors intuitively present a large transportation cost, that probably hinders the progress of \modelshort{}, as we have demonstrated its relationship to OT in \Cref{sec:method}.

To facilitate the workings of \modelshort{}, we follow a heuristic to transform the colors of the segmentation images. Specifically, on a small subset of the train dataset, we run an OT algorithm to compute a color correspondence that minimizes the color differences in terms of Sinkhorn distances between the real and segmentation images. The segmentation (target) datasets undergo this color conversion before they are fed into a diffusion model for training. During evaluation, when we compute MSEs, the images are converted to the original color space.

\paragraph{Privacy Protection} Color conversion requires considering both datasets jointly to compute a color mapping, and seems to betray the original purpose of \modelshort{} on protection of dataset privacy. We comment that the amount of leaked information is minimal: for example, to compute a color correspondence for the Maps dataset, we sampled only around 1000 pixels from the two datasets, to summarize the color composition information. \modelshort{} still conserve privacy at large.

\newpage
\subsection{Example-Guided Color Transfer}
\label{appendix:color-transfer}

\begin{figure*}[ht]
\begin{center}
\centerline{\includegraphics[width=\linewidth]{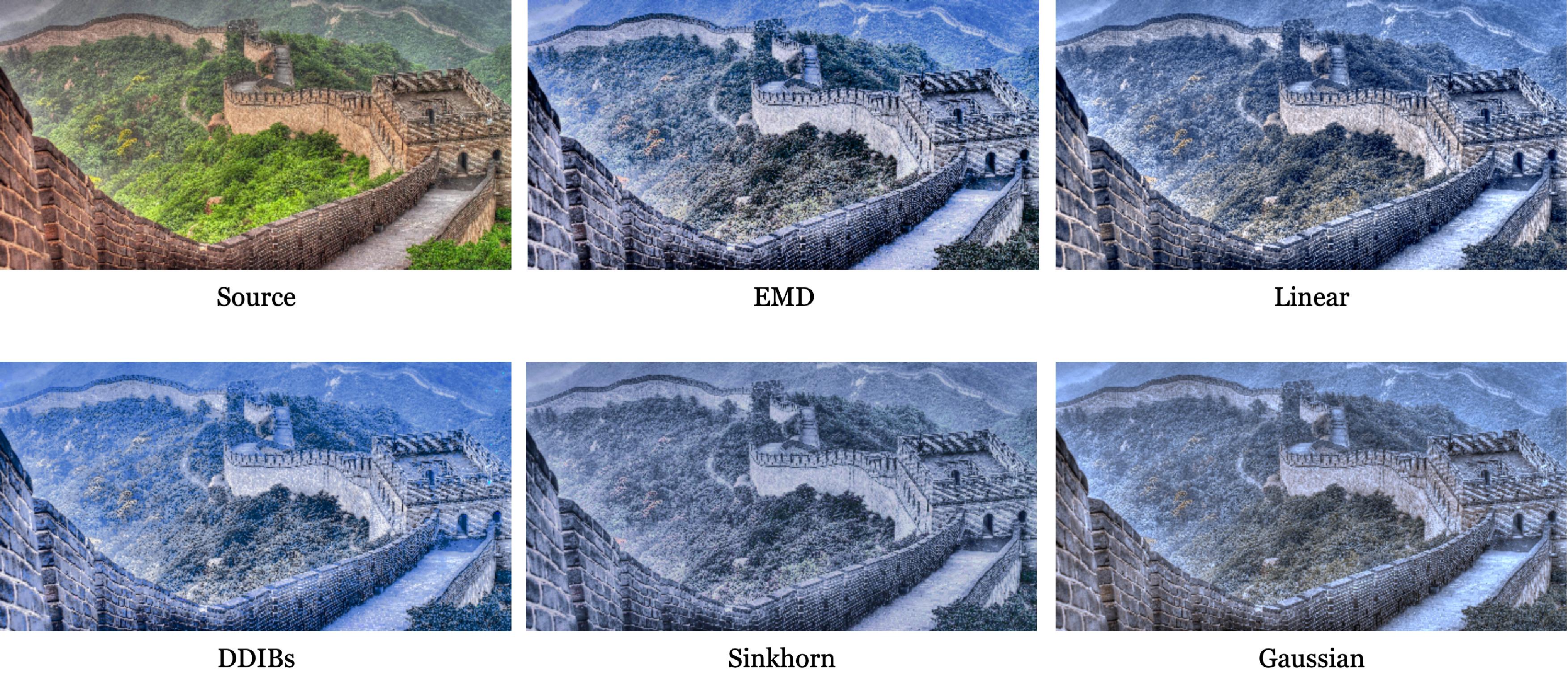}}
\vskip 0.2in
\centerline{\includegraphics[width=0.9\linewidth]{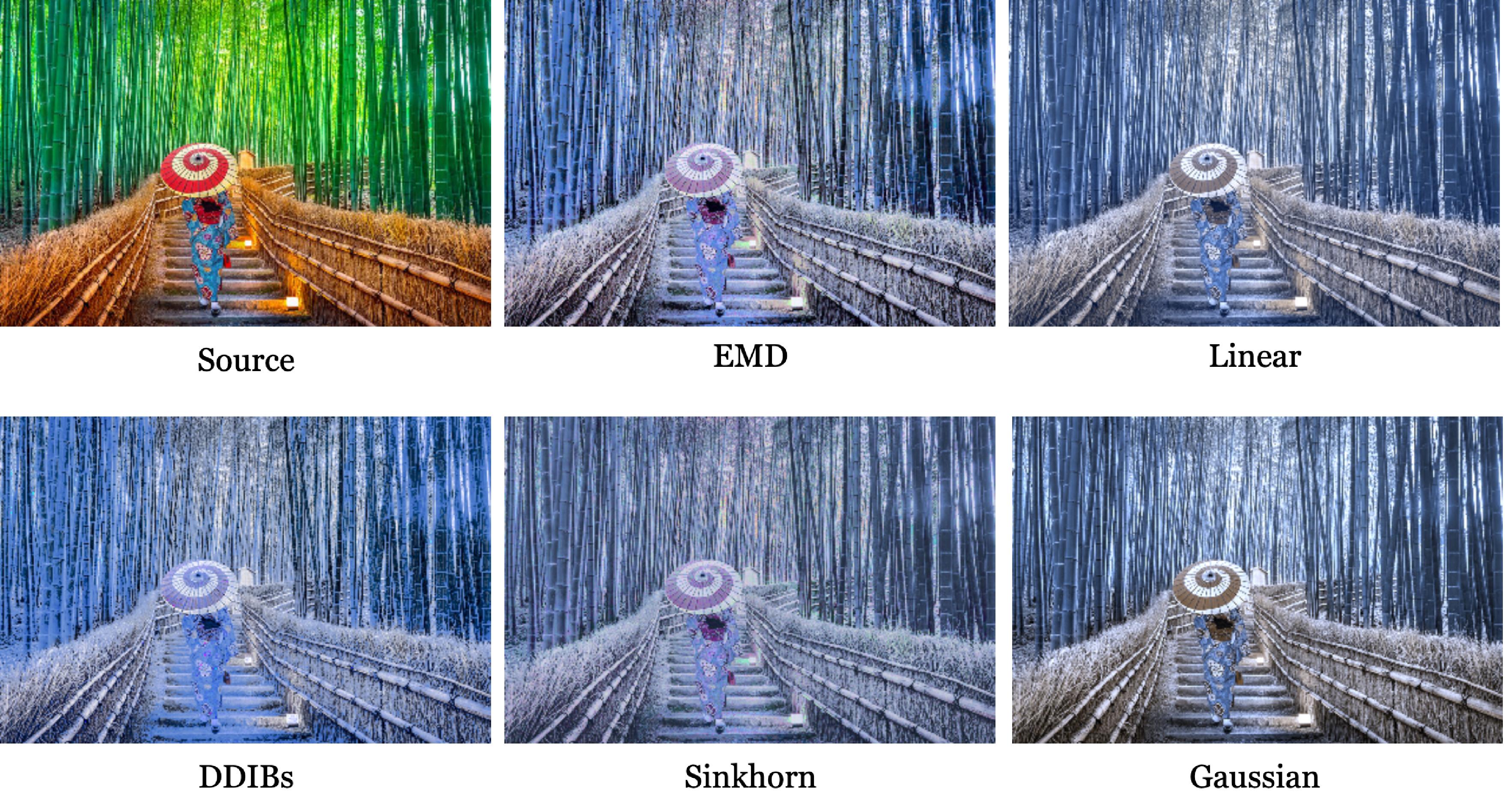}}
\caption{Full color transfer results on example images.}
\label{fig:color_translation_full_}
\end{center}
\vskip -0.2in
\end{figure*}

We present additional qualitative comparison between \modelshort{} and common OT methods, in \Cref{fig:color_translation_full_}.

\end{document}